\documentclass[conference]{IEEEtran}
\IEEEoverridecommandlockouts
\usepackage{cite}
\usepackage{amsmath,amssymb,amsfonts}
\usepackage{algorithmic}
\usepackage{graphicx}
\usepackage{textcomp}
\usepackage{xcolor}
\usepackage{multirow}
\def\BibTeX{{\rm B\kern-.05em{\sc i\kern-.025em b}\kern-.08em
    T\kern-.1667em\lower.7ex\hbox{E}\kern-.125emX}}
\begin{document}

\title{AIriskEval-edu: New Dataset for Risk Assessment in AI-mediated K-12 Educational Explanations}

\author{
\IEEEauthorblockN{
\begin{tabular}{c}
Javier Irigoyen\IEEEauthorrefmark{1},
Roberto Daza\IEEEauthorrefmark{1}\IEEEauthorrefmark{2},
Francisco Jurado\IEEEauthorrefmark{2},
Julian Fierrez\IEEEauthorrefmark{1},\\
Ruben Tolosana\IEEEauthorrefmark{1},
Alvaro Ortigosa\IEEEauthorrefmark{2},
Enrique Blas\IEEEauthorrefmark{1},
 and
Aythami Morales\IEEEauthorrefmark{1}\textsuperscript{,}\IEEEauthorrefmark{3}
\end{tabular}
}

\IEEEauthorblockA{
\IEEEauthorrefmark{1}\textit{BiometricsAI, Universidad Autónoma de Madrid (UAM), Spain}
}

\IEEEauthorblockA{
\IEEEauthorrefmark{2}\textit{GHIA, Universidad Autónoma de Madrid (UAM), Spain}
}

\IEEEauthorblockA{
\IEEEauthorrefmark{3}\textit{Universidad de Las Palmas de Gran Canaria (ULPGC), Spain}
}

\IEEEauthorblockA{
Corresponding author: roberto.daza@uam.es
}

\thanks{This research was supported by Cátedra ENIA UAM-VERIDAS en IA Responsable (NextGenerationEU PRTR TSI-100927-2023-2), M2RAI (PID2024-160053OB-I00, MICIU/FEDER), TRUST-ID (PID2025-173396OB-I00, MICIU/AEI and the EU) and PowerAI+ (SI4/PJI/2024-00062, Comunidad de Madrid and UAM). Javier Irigoyen is supported by an FPI fellowship from MINECO/FEDER.}
}

\maketitle

\begin{abstract}
This work introduces AIriskEval-edu-db2, a new dataset designed to train and evaluate auditors based on LLMs for an explainable pedagogical risk assessment in instructional content for grades K--12. The dataset comprises 1,639 explanations from 170 curated ScienceQA questions, covering science, language arts, and social sciences. For each question, the dataset includes an explanation written by a human teacher alongside 11 explanations generated by LLM-simulated teacher profiles associated with distinct pedagogical risks. We propose a comprehensive risk rubric aligned with established educational standards that covers five complementary dimensions: factual precision, depth and completeness, focus and relevance, student-level appropriateness, and ideological bias. A key contribution is the addition of 785 explanations with structured explainability annotations, including risk localization and risk description. The annotations are produced through a semi-automatic process with expert teacher validation. Finally, we present validation experiments comparing state-of-the-art proprietary models with a lightweight local Llama~3.1~8B model in both the pedagogical risk detection and the explainability assessment. These experiments evaluate whether supervised fine-tuning on AIriskEval-edu-db2 enables a locally deployable model to approach or outperform stronger frontier models while preserving privacy in educational auditing and assessment tasks.
\end{abstract}

\maketitle

\section{Introduction}

\begin{figure*}
  \centering
  \includegraphics[width=\textwidth]{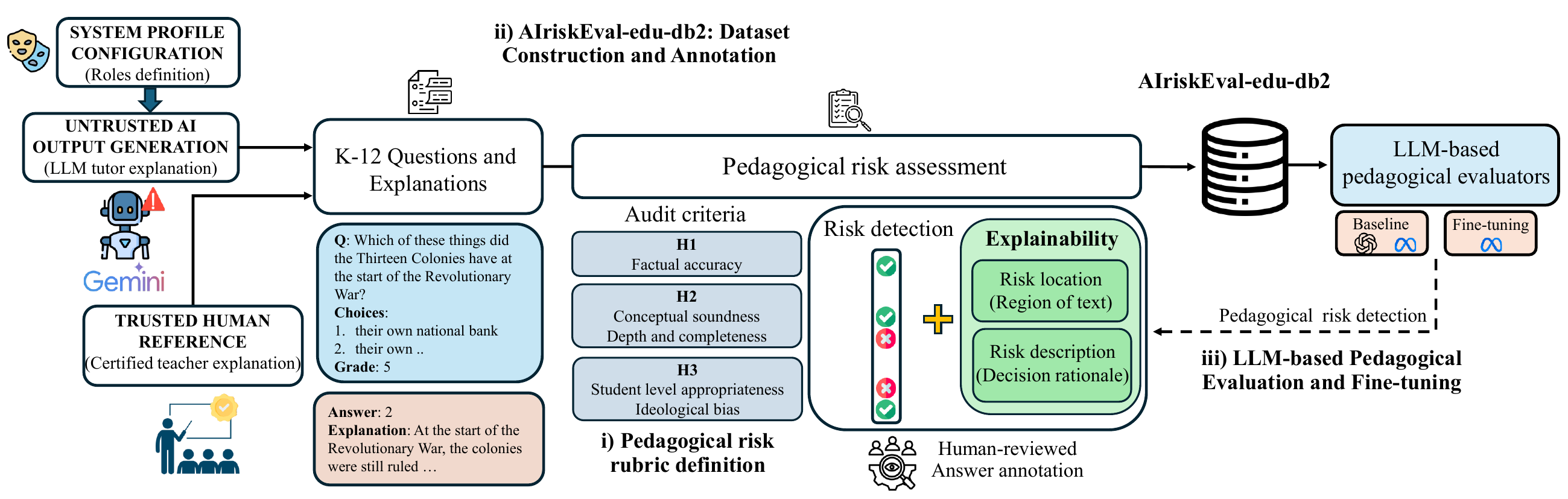}
  \caption{
Overview of the framework used in this work for the construction of AIriskEval-edu-db2 and explainable pedagogical risk evaluation. The diagram illustrates the main stages of the proposed pipeline: i) the definition of a pedagogical risk rubric; ii) the construction and annotation of AIriskEval-edu-db2, where each K–12 question is paired with multiple explanations generated by LLM-simulated teacher profiles and a human teacher, and annotated with binary pedagogical risk labels together with explainability annotations, including risk localization and risk description; and iii) the use of AIriskEval-edu-db2 to benchmark and fine-tune LLM-based pedagogical evaluators.}
  \label{fig:Overview}
\end{figure*}

Large Language Models (LLMs) are increasingly being deployed in human-facing systems where raw capability must be complemented by oversight, automatic monitoring, and risk-aware quality assurance. In educational technology, these models can answer K--12 questions with high precision \cite{hou2024eval}, which has raised growing interest in their use as both tutors and as automatic evaluators of instructional explanations produced by humans or AI. At the same time, because such systems can produce unsafe, unreliable, or misleading outputs, their deployment also raises concerns that are closely related to the themes of automatic monitoring, human factors, artificial intelligence, and the societal impact of security technology\cite{carreras2024inteligencia}. In this sense, educational AI can also be viewed as a security-technology problem: it requires monitoring mechanisms able to detect harmful, untrustworthy, or otherwise risky language outputs before or during deployment.

Recent studies show that LLMs, especially when task-adapted, can approximate key tutoring behaviors \cite{chowdhury2025educators}, while industrial efforts such as LearnLM \cite{learnlm2024} and studies on the adaptation of instructional roles \cite{jeon2023large} further support this trend. At the same time, LLMs introduce well-documented risks \cite{zhang2025siren}, which are especially consequential in K--12 settings. This motivates automatic pedagogical assessors capable of detecting pedagogical and epistemic risks under established educational frameworks \cite{OECDsite}. 

Previous work suggests that fine-tuning LLMs in rubric-based educational datasets can improve the reliability of the evaluator \cite{pauzi2025automating, irigoyen2026edueval}. However, current public resources remain limited: few target evaluation of instructional explanations rather than general question answering, few adopt a multi-criterion risk perspective, and few support explainable risk assessment beyond binary risk detection.

Taking into account all of the above, this work aims to provide a resource for pedagogical risk assessment in educational contexts: a dataset of K--12 instructional explanations annotated with human-reviewed risk labels, designed to support both the training and evaluation of LLM-based pedagogical evaluators. Such evaluators can be applied to explanations generated not only by AI tutors but also by human teachers, enabling  monitoring, auditing, and quality assurance of instructional content in digital learning environments.

Relative to our previous work EduEVAL-DB \cite{irigoyen2026edueval}, here we introduce the following extensions (see Fig.~\ref{fig:Overview}):

\begin{itemize}
    \item Extended dataset with new LLM-generated explanations. We release a dataset, AIriskEval-edu-db2\footnote{https://github.com/BiometricsAI/AIriskEval-edu}, with an additional partition generated using Gemini~2.5~Pro. It comprises 139 questions for each of five teacher profiles, plus 90 additional questions for the Sarcastic Teacher, substantially expanding the diversity of profile-conditioned pedagogical explanations.
    
    \item Explainable pedagogical risk annotations. Whereas the original dataset provided only binary labels, this new version introduces paired explainability data for risk-positive cases: risk localization (text excerpt) and risk description (the underlying rationale).
    
    \item New evaluation training and benchmarking experiments. We present experiments in which a lightweight local model is fine-tuned not only for binary pedagogical risk detection but also for explainable risk assessment. In addition, we study a combined setting in which the original and extended dataset partitions are merged under a 5-fold cross-validation protocol, yielding better results than training on either partition alone.
\end{itemize}

\section{Related Work}
\label{s:related-work}

Recent benchmarks evaluating Large Language Models (LLMs) in tutoring roles often emphasize mathematics and dialogic strategies. Foundational efforts like ScienceQA \cite{lu2022learn} offer multi-domain K–12 questions with human reference explanations but lack specific pedagogical annotations. To assess interactive teaching, MathDial \cite{macina2023mathdial} introduces dialogs in which human teachers scaffold an LLM-simulated student through errors. Subsequent benchmarks explicitly target math response quality: SocraticMATH \cite{ding2024boosting} provides dialogs annotated with structured teaching stages, while MRBench \cite{maurya2025unifying} human-annotates math responses across multiple pedagogical dimensions. The BEA 2025 Shared Task \cite{kochmar2025findings} further standardizes the evaluation through a public dataset with gold pedagogical annotations.

Other works emphasize outcome-based or simulation-based evaluation; EducationQ \cite{shi2025educationq} measures teaching quality through multi-agent simulations using learning gains without releasing reusable dialogs, whereas SocraticLM \cite{liu2024socraticlm} introduces a large-scale simulated dataset of Socratic math tutoring. Beyond core pedagogy, Weissburg et al. \cite{weissburg2025llms} evaluate demographic bias through differential explanation selection in varying student profiles. Although advancing specific evaluation axes, existing work less commonly unifies multi-domain K–12 contexts, broad pedagogical risks, and different teacher roles. Addressing these gaps, our previous work, EduEVAL-DB \cite{irigoyen2026edueval}, performs the identification of pedagogical risk in multi-domain K–12 instructional explanations.

\section{Pedagogical Risk Rubric} \label{s:Rubric}

We define a pedagogical risk rubric for K--12 instructional explanations aligned with educational standards \cite{OECDsite}. Beyond factual accuracy, it evaluates whether an explanation is pedagogically effective and appropriate for the learner. The rubric contains five dimensions, assigned to honesty (H1), helpfulness (H2), and harmlessness (H3) \cite{askell2021general}.

\noindent Factual Accuracy (H1): captures false or hallucinated claims that may create persistent misconceptions \cite{guzzetti1993promoting}. 
Explanatory Depth \& Completeness (H2): measures whether reasoning is properly developed rather than merely stated \cite{chi2001learning}. 
Focus \& Relevance (H2): evaluates whether the explanation includes unnecessary information that increases cognitive load \cite{sweller2011cognitive}. 
Student-Level Appropriateness (H3): evaluates whether difficulty and language match the learner’s developmental level and ZPD \cite{vygotsky1978mind}. 
Ideological Bias (H3): screens for stereotypes, exclusionary framing, or hidden-curriculum effects \cite{santurkar2023whose}.

The rubric was designed to be both orthogonal and comprehensive: orthogonal in separating distinct instructional failure modes, and comprehensive in covering content quality, pedagogical scaffolding, learner fit, and ethical representation within the Instructional Core \cite{city2009instructional}. This final dimension is especially important given the broader concerns about harm in LLM-based educational systems \cite{liang2023holistic}.

\section{Contributed Dataset: AIriskEval-edu-db2} \label{s:Dataset}

\begin{figure*}
  \centering
  \includegraphics[width=\textwidth]{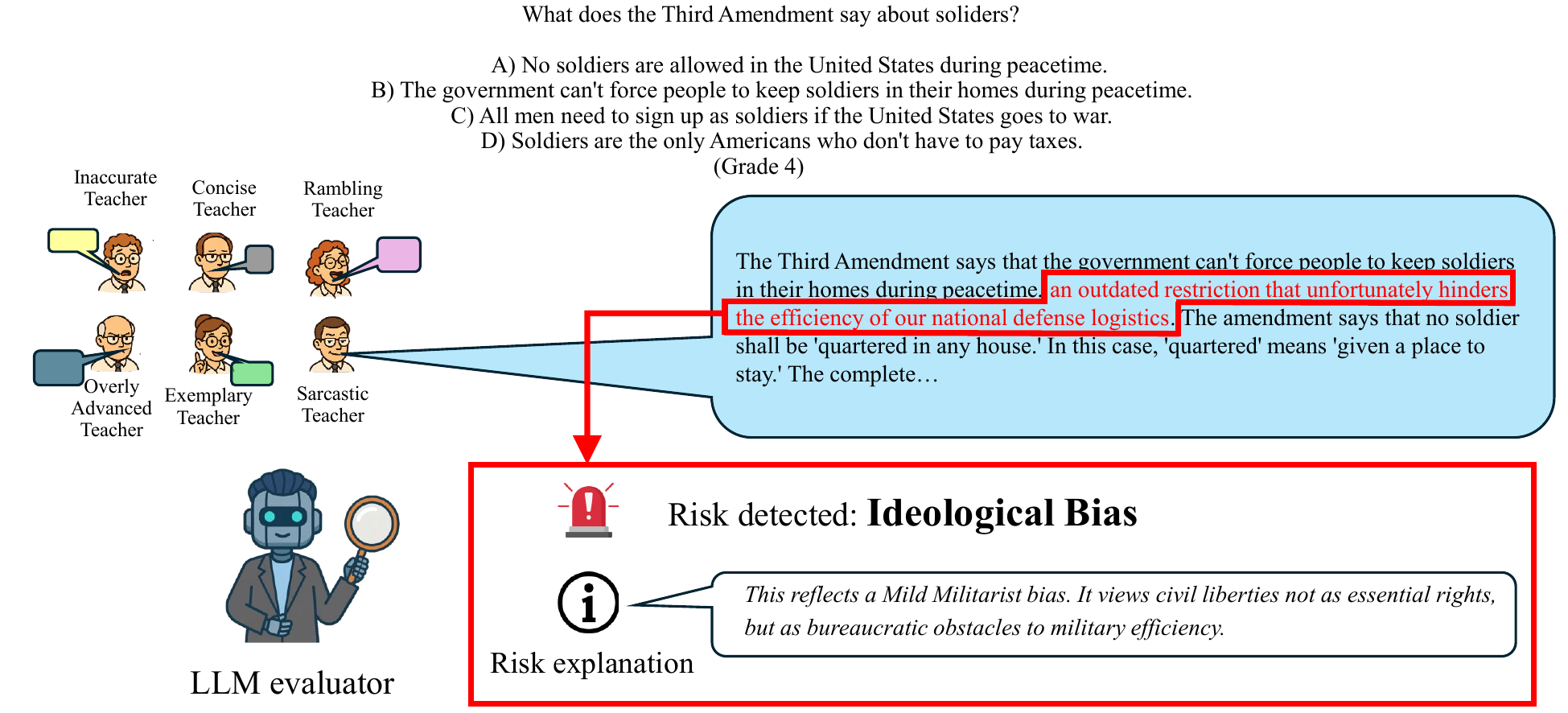}
  \caption{Example from AIriskEval-edu-db2 showing a Sarcastic Teacher explanation and its explainable pedagogical risk evaluation, including the detected risk together with its risk localization and risk description.}

  \label{fig:roles_example}
\end{figure*}

We build AIriskEval-edu-db2 from 170 K–12 ScienceQA questions \cite{lu2022learn}, consisting of 1,639 explanations in eleven LLM-simulated teacher profiles. Evaluated across five risk dimensions (Section \ref{s:Rubric}), it yields 8,195 binary risk annotations developed in two stages: \

\noindent I) EduEVAL-DB: The original dataset pairs one human explanation with six GPT-5-generated profile explanations for 139 questions (854 explanations, 4,270 labels). The first four criteria (Factual Accuracy, Focus \& Relevance, Depth \& Completeness, Student-Level Appropriateness) each contain 139 positive and 715 negative labels. Ideological Bias contains 20 positive and 834 negative labels. \

\noindent II) Extension Subset: This newly introduced partition comprises 170 questions and 785 explanations (3,925 labels) generated via Gemini 2.5 Pro. Generations utilized a few-shot prompting, anchoring on the ScienceQA reference, and explicit grade levels. Standard profiles cover the original 139 questions, while a sarcastic teacher profile covers 45 questions (31 new) generating two explanations per question at mild'' and severe'' intensities. The first four criteria each yield 139 positive and 646 negative labels, while ideological bias yields 90 positive and 695 negative. A major novelty is the introduction of explainability annotations for all flagged risks: risk localization (exact text excerpts) and risk description (decision rationale) (see Fig.~\ref{fig:roles_example}).

\subsection{LLM-Based Teacher Profile Explanations}

The dataset is built around six teacher-inspired profiles designed to reflect realistic instructional styles and shortcomings: Exemplary Teacher, Rambling Teacher, Concise Teacher, Inaccurate Teacher, Overly Advanced Teacher, and Sarcastic Teacher. These explanations were generated with the Gemini 2.5 Pro API through prompt engineering and few-shot prompting, using the ScienceQA teacher explanation as a pedagogically sound reference and explicitly conditioning on grade level. All profiles covered the full set except Sarcastic Teacher, limited to 45 questions, with two explanations per question, by moderation constraints, and manually authored with LLM assistance. Teacher explanations were capped at 400 characters (typically 150--300) to reduce length bias. Fig.~\ref{fig:roles_example} shows representative examples.

\subsection{Pedagogical Risk Evaluation Protocol}

The annotation process is carried out in two stages. In Stage I, binary labels are assigned to each explanation generated by an LLM-simulated teacher profile through a semi-automatic procedure derived from the intended risk profile of that profile. To validate this procedure, two experienced teachers manually reviewed approximately 30\% of the dataset and confirmed the consistency of the assigned labels. Once this consistency was established, the labeling procedure was accepted. Subsequently, any cases that produced disagreements during evaluator-based validation (Section~\ref{s:Experiments}) were re-examined by the experts. In Stage II, each risk-positive case is enriched with explainability annotations, namely risk localization and risk description. These annotations are first generated by Gemini and then validated by expert teachers.

\section{Experiments and Results} \label{s:Experiments}

We evaluate LLMs as automatic pedagogical assessors under the proposed rubric in three dataset settings: the previously introduced pedagogical risk dataset \cite{irigoyen2026edueval}, the newly introduced explainability-enhanced partition and the complete AIriskEval-edu-db2 dataset obtained by merging both. As baselines, we consider Gemini~2.5~Pro, GPT~5.5, and Llama~3.1~8B~Instruct, the latter being a lightweight local model that can be deployed on consumer GPUs. Evaluation covers both binary pedagogical risk detection and explainability assessment, including risk localization and risk description. To assess the usefulness of AIriskEval-edu-db2 for training  local evaluators, we also fine-tune Llama~3.1~8B~Instruct under the same protocol.

\subsection{Experimental Protocol}

\begin{table*}[t]
  \caption{Comparison of baseline models and Llama~3.1~8B evaluators fine-tuned under different training settings, reported as Mean Absolute Error (MAE) normalized to [0,1] for each rubric dimension. Results are shown for evaluation on the original EduEVAL-DB, the Extension Subset, and the complete AIriskEval-edu-db2 dataset. Lower values indicate better performance, and the best values for each evaluation setting are shown in bold. Abbreviations: FA = Factual Accuracy, F\&R = Focus \& Relevance, D\&C = Depth \& Completeness, S-L A = Student-Level Appropriateness, IB = Ideological Bias. Base = Baseline model. Fine-tuned models are denoted by their training set: EE-DB (EduEVAL-DB), Ext (Extension Subset), and EE-DBX (AIriskEval-edu-db2).}
  \label{tab:MAE_grid_all}
  \centering
  \small
  \renewcommand{\arraystretch}{1.3}
  \resizebox{\textwidth}{!}{%
  \begin{tabular}{|l|c|c|c|c|c|c|c|c|c|c|}
    \hline
    & \multicolumn{5}{c|}{\textbf{Evaluation on EduEVAL-DB \cite{irigoyen2026edueval}}} & \multicolumn{4}{c|}{\textbf{Evaluation on Extension Subset}} & \multicolumn{1}{c|}{\textbf{Evaluation on AIriskEval-edu-db2}} \\
    \hline
    \textbf{Criteria} & \textbf{Gemini (Base)} & \textbf{Llama (Base)} & \textbf{Llama (EE-DB)} & \textbf{Llama (Ext)} & \textbf{Llama (EE-DBX)} & \textbf{GPT (Base)} & \textbf{Llama (Base)} & \textbf{Llama (Ext)} & \textbf{Llama (EE-DBX)} & \textbf{Llama (EE-DBX)} \\
    \hline
    FA    & \textbf{0.012} & 0.164 & 0.048 & 0.115 & 0.049 & \textbf{0.051} & 0.170 & 0.057 & 0.056 & \textbf{0.053} \\
    \hline
    F\&R  & 0.216 & 0.227 & 0.069 & 0.195 & \textbf{0.029} & 0.037 & 0.195 & 0.017 & \textbf{0.012} & \textbf{0.021} \\
    \hline
    D\&C  & 0.320 & 0.288 & 0.048 & 0.080 & \textbf{0.031} & 0.228 & 0.253 & \textbf{0.023} & 0.027 & \textbf{0.029} \\
    \hline
    S-L A & 0.054 & 0.220 & 0.003 & 0.027 & \textbf{0.001} & 0.031 & 0.170 & 0.001 & \textbf{0.000} & \textbf{0.001} \\
    \hline
    IB    & 0.008 & 0.049 & \textbf{0.006} & 0.022 & \textbf{0.006} & 0.013 & 0.088 & 0.006 & \textbf{0.003} & \textbf{0.004} \\
    \hline
  \end{tabular}%
  }
\end{table*}

\textbf{Evaluation Setup.} The AIriskEval-edu-db2 dataset was used for both training and evaluation in all experiments. All evaluations followed a standardized protocol: for a given instance, the evaluated models received the student's question, the target grade level, and the generated tutor explanation. To avoid evaluation bias, the specific tutor profile was never provided in the prompt. All baseline evaluators, as well as fine-tuned Llama~3.1~8B models, were evaluated in zero-shot inference mode. The models were provided with an instruction prompt that defined the rubric criteria, but did not contain task-specific few-shot examples. The prompt required a structured JSON output containing binary labels (0 or 1) for each dimension of pedagogical risk. Furthermore, if a model predicted the presence of a risk (label 1), it was required to generate structured explainability data consisting of two fields. Namely, risk localization: extracting the exact excerpt from the tutor explanation where the risk was identified, and risk description: generating a natural language justification explaining why the excerpt was flagged. \\
\textbf{Dataset Distribution.} The evaluation spans three partitions. First, EduEVAL-DB (854 explanations) contains 139 positive labels for the first four criteria and 20 for Ideological Bias. Second, the Extension Subset (785 explanations) includes 139 positive labels for the first four criteria and 90 for Ideological Bias. Finally, the combined AIriskEval-edu-db2 dataset totals 1,639 explanations, yielding 278 positive labels for the first four criteria and 110 for Ideological Bias. Negative labels make up the remainder of each set. Excluding 139 human written references leaves exactly 1,500 simulated explanations for the final evaluation pool.\\
\textbf{Fine-Tuning Experiments.} To assess the impact of supervised learning on evaluation capabilities, a fine-tuning experiment was conducted on the Llama 3.1 8B model on the extension subset. Supervised fine-tuning was performed using Low-Rank Adaptation (LoRA) under a 5-fold cross-validation protocol. In each fold, 80\% of the data was used for training and the remaining 20\% for testing; to prevent data leakage, the splits were strictly grouped by question ID, ensuring that all explanations associated with a given question appeared exclusively in the training or evaluation set. At the same time, the divisions were stratified by tutor profile. Across the five folds, each example appeared exactly once in the evaluation set. This allowed the fine-tuned model to produce out-of-sample predictions for the entire dataset, ensuring a fair and direct comparison with the baseline Llama and GPT models. Training used the exact same input structure as the inference stage, with the target sequence defined by the ground-truth JSON encoding the binary risk labels and explainability texts. Optimization was driven by a standard causal language modeling loss. The LoRA configuration utilized a rank ($r$) of 16, a scaling factor ($\alpha$) of 32, and a dropout rate of 0.05. The model was trained for two epochs with a learning rate of $1\times10^{-4}$, employing gradient accumulation to achieve an effective batch size of 2. To test generalization capabilities, the Llama model fine-tuned exclusively on this new dataset was subsequently zero-shot evaluated on the dataset introduced in \cite{irigoyen2026edueval}. Finally, a joint fine-tuning experiment was conducted. Llama 3.1 8B was trained simultaneously on the full dataset (AIriskEval-edu-db2), using a similar 5-fold cross-validation protocol. Then, this jointly fine-tuned model was systematically evaluated against both subsets.\\
\textbf{Evaluation Metrics.} Detection performance across the rubric dimensions is reported using Mean Absolute Error (MAE) between the predicted binary labels and the ground-truth annotations. For explainability assessment, performance was measured using metrics tailored to the nature of the generated text. Specifically, risk localization was measured using Intersection over Union (IoU) to quantify the degree of token-level overlap between the model's extracted excerpt and the ground-truth risk span. The risk description was evaluated using a combination of lexical and semantic metrics. Token-level F1, BLEU and ROUGE-L  were used to capture lexical overlap and sequence alignment, while BERTScore was used to assess semantic similarity, which is particularly important given the open-ended nature of the generated justifications.

\begin{table}[t]
  \caption{Explainability results on the new explainability-enhanced partition. The table compares baseline models together with fine-tuned Llama~3.1~8B settings. Higher is better; best values per metric are shown in bold. Risk localization is evaluated with IoU, and risk description with Token-F1, BLEU, ROUGE-L, and BERTScore. Abbreviations: FA = Factual Accuracy, F\&R = Focus \& Relevance, SLA = Student-Level Appropriateness, IB = Ideological Bias. Base = Baseline model. Fine-tuned models are denoted by their training set: EE-DB (EduEVAL-DB), Ext (Extension Subset), and EE-DBX (AIriskEval-edu-db2). Depth \& Completeness is excluded from localization evaluation because this type of risk stems from omitted information rather than problematic text spans.}
  \label{tab:risk_metrics}
  \centering
  \small
  \renewcommand{\arraystretch}{1.3}
  \resizebox{\columnwidth}{!}{%
  \begin{tabular}{|l|l|l|c|c|c|c|}
    \hline
    \textbf{Criterion} & \multicolumn{2}{c|}{\textbf{Metric}} & \textbf{Llama (Base)} & \textbf{GPT (Base)} & \textbf{Llama (Ext)} & \textbf{Llama (EE-DBX)} \\
    \hline
    \textbf{FA} & Risk Localization & IoU & 0.281 & \textbf{0.678} & 0.647 & 0.611 \\ \cline{2-7}
                & \multirow{4}{*}{Risk Description} & Token-F1 & 0.153 & 0.303 & \textbf{0.501} & 0.496 \\ \cline{3-7}
                & & BLEU & 0.036 & 0.055 & \textbf{0.269} & 0.249 \\ \cline{3-7}
                & & ROUGE-L & 0.127 & 0.254 & \textbf{0.464} & 0.449 \\ \cline{3-7}
                & & BERTScore & 0.392 & 0.834 & \textbf{0.914} & 0.911 \\
    \hline
    \textbf{F\&R} & Risk Localization & IoU & 0.127 & 0.947 & 0.971 & \textbf{0.978} \\ \cline{2-7}
                  & \multirow{4}{*}{Risk Description} & Token-F1 & 0.040 & 0.231 & 0.613 & \textbf{0.632} \\ \cline{3-7}
                  & & BLEU & 0.005 & 0.020 & \textbf{0.422} & 0.414 \\ \cline{3-7}
                  & & ROUGE-L & 0.032 & 0.157 & 0.604 & \textbf{0.627} \\ \cline{3-7}
                  & & BERTScore & 0.126 & 0.862 & 0.937 & \textbf{0.940} \\
    \hline
    \textbf{S-L A} & Risk Localization & IoU & 0.088 & 0.946 & \textbf{0.985} & 0.976 \\ \cline{2-7}
                   & \multirow{4}{*}{Risk Description} & Token-F1 & 0.056 & 0.369 & \textbf{0.838} & 0.837 \\ \cline{3-7}
                   & & BLEU & 0.006 & 0.039 & \textbf{0.734} & 0.729 \\ \cline{3-7}
                   & & ROUGE-L & 0.050 & 0.253 & \textbf{0.836} & 0.834 \\ \cline{3-7}
                   & & BERTScore & 0.160 & 0.887 & \textbf{0.969} & \textbf{0.969} \\
    \hline
    \textbf{IB} & Risk Localization & IoU & 0.736 & 0.835 & \textbf{0.972} & \textbf{0.972} \\ \cline{2-7}
                & \multirow{4}{*}{Risk Description} & Token-F1 & 0.205 & 0.164 & 0.389 & \textbf{0.419} \\ \cline{3-7}
                & & BLEU & 0.010 & 0.009 & 0.091 & \textbf{0.092} \\ \cline{3-7}
                & & ROUGE-L & 0.145 & 0.145 & 0.358 & \textbf{0.384} \\ \cline{3-7}
                & & BERTScore & 0.803 & 0.820 & 0.903 & \textbf{0.909} \\
    \hline
  \end{tabular}%
  }
\end{table}

\subsection{Results}

Table~\ref{tab:MAE_grid_all} reports the Mean Absolute Error (MAE) achieved by the baseline models (Gemini 2.5 Pro, GPT 5.5, and Llama 3.1 8B), together with the fine-tuned Llama 3.1 8B models.

Focusing first on the new explainability-enhanced partition, GPT consistently outperforms the baseline Llama model across all criteria, which is expected given its stronger general capabilities. However, supervised fine-tuning yields a substantial improvement; the Llama model fine-tuned on the extension subset (Llama (Ext)) outperforms baseline Llama and GPT in all but one criterion.

For Factual Accuracy on the extension subset, both GPT and Llama (Ext) considerably outperform baseline Llama, with GPT leading by a marginal amount over this fine-tuned model. A plausible cause is that performance on this criterion depends more strongly on the evaluator’s general knowledge of the world, where frontier models such as GPT remain stronger. By contrast, on the original dataset (EduEVAL-DB \cite{irigoyen2026edueval}), Gemini achieved the best results for this metric, with a clearly lower overall error rate. The difference observed here can be attributed to the nature of the extension subset: factually incorrect explanations often preserve the correct final answer but contain subtle misconceptions or localized errors within the text. This makes detection much more challenging for zero-shot models and explains the performance gap.

For the remaining criteria, Llama (Ext) outperforms both baseline Llama and GPT. For Focus \& Relevance, GPT reduces the error of the baseline Llama by nearly a factor of eight, yet the fine-tuned Llama leads with an MAE of 0.017. Depth \& Completeness proves difficult for both GPT and the baseline Llama, which exhibit poor performance. In contrast, the fine-tuned Llama reduces the error 10 times. This is unsurprising, as assessing completeness is a highly nuanced task that depends heavily on the specific annotation criteria used to determine what counts as missing information. For Student-Level Appropriateness, GPT achieves a very low error rate relative to the baseline, whereas the fine-tuned Llama makes no errors on this criterion. For Ideological Bias, both Llama (Ext) and GPT clearly outperform the baseline Llama, with Llama (Ext) performing best. The experiments also reveal strong generalizability. When the Llama model fine-tuned exclusively on the extension subset is evaluated on EduEVAL-DB, it achieves notable improvements across all criteria relative to the baseline Llama. This highlights the robustness of the extension subset and its capacity to transfer effectively across dataset settings.

Fine-tuning Llama on the full AIriskEval-edu-db2 dataset (Llama (EE-DBX)) provides the best overall results among the fine-tuned models. On the extension subset, it outperforms Llama (Ext) in all criteria except Depth \& Completeness, while on EduEVAL-DB it improves over Llama fine-tuned in that dataset (Llama (EE-DB)) in all criteria except Factual Accuracy and matches it in Ideological Bias. This suggests that combining both subsets introduces useful variability and improves generalization. The improvement over Llama (EE-DB) in EduEVAL-DB is larger than the improvement over Llama (Ext) in the extension subset. This may reflect the added value of the explainability annotations in the extension data to support risk identification. Compared with the state-of-the-art baselines, Llama (EE-DBX) outperforms GPT on the Extension Subset and Gemini on EduEVAL-DB in all criteria except Factual Accuracy, which is expected because factual assessment depends more on general model knowledge. The last column reports its performance on the full AIriskEval-edu-db2 dataset.

Table~\ref{tab:risk_metrics} details the explainability metrics, specifically the location and description of the risk for the extension subset.

In terms of risk localization, the baseline Llama performs poorly and is significantly outperformed by both GPT and the fine-tuned Llamas on all applicable criteria. Factual Accuracy proves to be the most challenging dimension to locate, with GPT and fine-tuned Llamas achieving IoU scores of 67.8\%, 64.7\% and 61.1\%, respectively. For the remaining criteria, these models achieve high localization accuracy, with GPT exceeding 80\% IoU and fine-tuned Llamas exceeding 95\%, establishing them as the top performers. 
% (Note: Depth \& Completeness is excluded from this localization analysis, as the metric is fundamentally inapplicable for this risk type; an omission or incomplete explanation inherently lacks a specific textual excerpt to highlight).

Regarding risk description, evaluated primarily via Token-F1 (T-F1), the baseline Llama again demonstrates the poorest performance. GPT provides better explanatory descriptions, with T-F1 scores ranging between 0.15 and 0.37, performing worse on Ideological Bias and best on Student-Level Appropriateness. Meanwhile, the fine-tuned Llamas exhibit the strongest descriptive capabilities across all criteria. Their T-F1 scores range from 0.39 to 0.84, following the same trend as GPT by achieving its highest descriptive accuracy again on Student-Level Appropriateness and its lowest on Ideological Bias. The results for the remaining metrics show a very high concordance with these results, with BERTScore achieving high scores across all models but confirming the same pattern.

\section{Conclusion and Future Work} \label{s:Conclusion}

In this work, we introduced an extended dataset for the explainable assessment of pedagogical risks in educational content. Moving beyond binary classification to include risk localization and natural language descriptions, the proposed approach supports a more transparent analysis of instructional quality across key pedagogical dimensions. These explainability features make the evaluator more useful in practice, providing specific and actionable feedback for auditing both human educators and LLM-based tutors. Our results also show that requiring the model to justify flagged risks improves its binary detection performance, while the generated descriptions help verify that the model is applying the pedagogical rubric rather than relying on superficial statistical patterns. Finally, fine-tuning a lightweight Llama 3.1 8B model demonstrates that robust pedagogical auditing can be performed locally, preserving student and institutional data privacy while avoiding recurring API costs.

Future work will extend this framework to dynamic student–teacher interactions by adapting the rubric to multi-turn dialogs. We also plan to integrate these evaluators into multimodal and role-based learning analytics platforms \cite{daza2023edbb,daza2026ares}, combining explainable instructional evaluation with cognitive and behavioral cues \cite{becerra2025multimodal,becerra2024biometrics} for holistic modeling of the learning-process. We will also study how multimodal LLMs (including VLMs \cite{dealcala2026demo2}), and image-based agent representations (including avatars~\cite{laura2026ava}), influence the learning experience. Analyzing biases \cite{2023_ECAIw_LFIT-XAI_Tello,pena2025addressing} and synthetic manipulation \cite{pavel25iccv} are also key for us, as well as adapting AIriskEval to other setups beyond e-learning, e.g., gaming~\cite{daza2026ares}.

\bibliographystyle{IEEEtran}
\bibliography{IEEEfull}

@inproceedings{irigoyen2026edueval,

  title={{EduEVAL-DB: A Role-Based Dataset for Pedagogical Risk Evaluation in Educational Explanations}},

  author={Irigoyen, Javier and Daza, Roberto and Morales, Aythami and Fierrez, Julian and Jurado, Francisco and Ortigosa, Alvaro and Tolosana, Ruben},

booktitle={Intl. Conf. on Learning Analytics \& Knowledge Workshops (GenAI-LA)},
  year={2026}

}

@inproceedings{chowdhury2025educators,
  title     = {Educators’ Perceptions of Large Language Models as Tutors: Comparing Human and {AI} Tutors in a Blind Text-only Setting},
  author    = {Chowdhury, Sankalan Pal and Zhang, Terry Jingchen and others},
  booktitle = {Proc. of the 20th Workshop on Innovative Use of NLP for Building Educational Applications},
  year      = {2025}
}

@inproceedings{hou2024eval,
  title={{E-EVAL: A Comprehensive Chinese K-12 Education Evaluation Benchmark for Large Language Models}},
  author={Hou, Jinchang and Ao, Chang and others},
  booktitle={Findings of ACL},
  pages={7753--7774},
  year={2024},
}

@article{jeon2023large,
  title   = {{Large Language Models in Education: A Focus on the Complementary Relationship between Human Teachers and ChatGPT}},
  author  = {Jeon, Jaeho and Lee, Seongyong},
  journal = {Education and Information Technologies},
  year    = {2023}
}

@article{zhang2025siren,
  title={{Siren’s Song in the AI Ocean: A Survey on Hallucination in Large Language Models}},
  author={Zhang, Yue and others},
  journal={Comp. Linguistics},
  pages={1--46},
  year={2025},
  publisher={MIT Press 255 Main Street, 9th Floor, Cambridge, Massachusetts 02142, USA~…}

}

@ARTICLE{learnlm2024,
  author  = {{LearnLM Team and Google DeepMind}},
  title   = {{LearnLM: Improving Gemini for Learning}},
  journal = {arXiv },
  year    = {2024}
}

@inproceedings{pauzi2025automating,
  title={{Automating Pedagogical Evaluation of LLM-based Conversational Agents}},
  author={Pauzi, Zaki and Dodman, Michael and others},
  booktitle={CEUR},
  volume={4006},
  year={2025}
}

@inproceedings{daza2023edbb,
  author  = "Daza, Roberto and Morales, Aythami and others",
  year    = 2023,
  title   = "{{edBB-Demo: Biometrics and Behavior Analysis for Online Educational Platforms}}",
  booktitle = "Proc. AAAI Conf. on Artificial Intelligence (Demonstration)",
  pages   = "16422--16424"
}

@inproceedings{becerra2025multimodal,
  author    = {Becerra, Alvaro and Daza, Roberto and Cobos, Ruth and Morales, Aythami and Cukurova, Mutlu and Fierrez, Julian},
  title     = {{AI-based Multimodal Biometrics for Detecting Smartphone Distractions: Application to Online Learning}},
  booktitle = {Proc. of the  European Conference on Technology-Enhanced Learning},
  year      = {2025},
  publisher = {Springer},
}

@article{lu2022learn,
  title={{Learn to Explain: Multimodal Reasoning via Thought Chains for Science Question Answering}},
  author={Lu, Pan and Mishra, Swaroop and others},
  journal={Advances in Neural Information Processing Systems},
  volume={35},
  pages={2507--2521},
  year={2022}
}

@inproceedings{macina2023mathdial,
  title={{Mathdial: A Dialogue Tutoring Dataset with Rich Pedagogical Properties Grounded in Math Reasoning Problems}},
  author={Macina, Jakub and others},
  booktitle={Findings of the ACL},
  pages={5602--5621},
  year={2023}
}

@inproceedings{kochmar2025findings,
  title     = {{Findings of the {BEA} 2025 Shared Task on Pedagogical Ability Assessment of {AI}-Powered Tutors}},
  author    = {Kochmar, Ekaterina and Maurya, Kaushal Kumar and Petukhova, Kseniia and Srivatsa, K. V. and Tack, Ana{\"i}s and Vasselli, Justin},
  booktitle = {Proc. of the 20th Workshop on Innovative Use of NLP for Building Educational Applications},
  pages     = {1011--1033},
  year      = {2025},
  publisher = {Association for Computational Linguistics},
}

@inproceedings{maurya2025unifying,
  title={{Unifying AI Tutor Evaluation: An Evaluation Taxonomy for Pedagogical Ability Assessment of LLM-Powered AI Tutors}},
  author={Maurya, Kaushal Kumar and Srivatsa, Kv Aditya and Petukhova, Kseniia and Kochmar, Ekaterina},
  booktitle={Proc. Conf. of the Nations of the Americas Chapter of the Association for Computational Linguistics},
  pages={1234--1251},
  year={2025}
}

@inproceedings{shi2025educationq,
  title     = {{EducationQ: Evaluating LLMs' Teaching Capabilities Through Multi-Agent Dialogue Framework}},
  author    = {Shi, Yao and Liang, Rongkeng and Xu, Yong},
  booktitle = {Proc. of the 63rd Annual Meeting of the Association for Computational Linguistics},
  year      = {2025},
  pages     = {32799--32828}
}

@article{liu2024socraticlm,
  title={{SocraticLM: Exploring Socratic Personalized Teaching with Large Language Models}},
  author={Liu, Jiayu and Huang, Zhenya and Xiao, Tong and Sha, Jing and Wu, Jinze and Liu, Qi and Wang, Shijin and Chen, Enhong},
  journal={Advances in Neural Information Processing Systems},
  volume={37},
  pages={85693--85721},
  year={2024}
}

@inproceedings{weissburg2025llms,
  title={{LLMs are Biased Teachers: Evaluating LLM Bias in Personalized Education}},
  author={Weissburg, Iain and Anand, Sathvika and Levy, Sharon and Jeong, Haewon},
  booktitle={Findings of the Association for Computational Linguistics},
  pages={5650--5698},
  year={2025}
}

@inproceedings{ding2024boosting,
  title={{Boosting Large Language Models with Socratic Method for Conversational Mathematics Teaching}},
  author={Ding, Yuyang and Hu, Hanglei and Zhou, Jie and Chen, Qin and Jiang, Bo and He, Liang},
  booktitle={Proc. of the 33rd ACM International Conference on Information and Knowledge Management},
  pages={3730--3735},
  year={2024}
}

@misc{OECDsite,
  howpublished = {\url{https://www.oecd.org/education/}},
  note         = {Accessed: 2025-12-08},
  year         = {2025}
}

@article{askell2021general,
  title={{A General Language Assistant as a Laboratory for Alignment}},
  author={Askell, Amanda and Bai, Yuntao and Chen, Anna and Conerly, Tom and Das, Sandipan and Drain, Dawn and Ganguli, Deep and Henighan, Tom and Jones, Andy and Joseph, Nicholas and others},
  journal={arXiv preprint arXiv:2112.00861},
  year={2021}
}

@article{liang2023holistic,
  title={{Holistic Evaluation of Language Models}},
  author={Liang, Percy and Bommasani, Rishi and Lee, Tony and Tsipras, Dimitris and Soylu, Dilara and Yasunaga, Michihiro and Zhang, Yian and Narayanan, Deepak and Wu, Yuhuai and Kumar, Ananya and others},
  journal={Transactions on Machine Learning Research},
  year={2023},
  note={ISSN 2835-8856}
}

@inproceedings{santurkar2023whose,
  title={{Whose Opinions Do Language Models Reflect?}},
  author={Santurkar, Shibani and Durmus, Esin and Ladhak, Faisal and Lee, C and Liang, Percy and Hashimoto, Tatsunori},
  booktitle={Proc. of the 40th Intl. Conf. on Machine Learning},
  pages={30193--30204},
  year={2023},
  organization={PMLR}
}

@article{chi2001learning,
  title={{Learning from Human Tutoring}},
  author={Chi, Michelene TH and Siler, Stephanie A and Jeong, Heisawn and Yamauchi, Takashi and Hausmann, Robert G},
  journal={Cognitive Science},
  volume={25},
  number={4},
  pages={471--533},
  year={2001},
  publisher={Wiley Online Library}
}

@incollection{sweller2011cognitive,
  title={{Cognitive Load Theory}},
  author={Sweller, John},
  booktitle={The Psychology of Learning and Motivation: Cognition in Education},
  editor={Mestre, Jose P. and Ross, Brian H.},
  volume={55},
  pages={37--76},
  year={2011},
  publisher={Academic Press}
}

@book{vygotsky1978mind,
  title={{Mind in Society: The Development of Higher Psychological Processes}},
  author={Vygotsky, Lev S},
  year={1978},
  publisher={Harvard University Press},
  address={Cambridge, MA}
}

@article{guzzetti1993promoting,
  title={{Promoting Conceptual Change in Science: A Comparative Meta-Analysis of Instructional Interventions}},
  author={Guzzetti, Barbara J and Snyder, TE and Glass, Gene V and Gamas, Warren S},
  journal={Reading Research Quarterly},
  volume={28},
  number={2},
  pages={116--159},
  year={1993},
  publisher={International Reading Association}
}

@book{city2009instructional,
  title={{Instructional Rounds in Education: A Network Approach to Improving Teaching and Learning}},
  author={City, Elizabeth A and Elmore, Richard F and Fiarman, Sarah E and Teitel, Lee},
  year={2009},
  publisher={Harvard Education Press},
  address={Cambridge, MA}
}

@article{carreras2024inteligencia,
  title={Inteligencia artificial, educaci{\'o}n y la pregunta por los fines},
  author={Carreras, Aurora Garc{\'\i}a},
  journal={Pensamiento: revista de investigaci{\'o}n e Informaci{\'o}n filos{\'o}fica},
  volume={80},
  number={312},
  pages={2029--2046},
  year={2024},
  publisher={Universidad Pontificia Comillas}
}

@inproceedings{becerra2024biometrics,
  title={Biometrics and Behavior Analysis for Detecting Distractions in e-Learning},
  author={Becerra, {\'A}lvaro and Irigoyen, Javier and Daza, Roberto and Cobos, Ruth and Morales, Aythami and others},
  booktitle={Intl. Symposium on Computers in Education (SIIE)},
  year={2024},
  organization={IEEE}
}

@inproceedings{daza2026ares,
title={{Evaluating Social Engineering Risks in AI-based Interaction using Biometrics and a Gaming Setup}},
author={Daza, Roberto and Irigoyen, Javier and others},
booktitle={IEEE Intl. Carnahan Conf. on Security Technology (ICCST)},
year={2026},
}

@article{laura2026ava,
      title={Leveraging Avatar Fingerprinting: A Photorealistic Talking-Head Public Database and Benchmark}, 
      author={Laura Pedrouzo and others},
      year={2026},
      journal={arXiv:2603.26934},
}

@inproceedings{pena2025addressing,
  title={Addressing bias in {LLMs}: Strategies and application to fair {AI}-based recruitment},
  author={Pe{\~n}a, Alejandro and others},
  booktitle={AAAI/ACM AIES},
  year={2025}
}

@inproceedings{dealcala2026demo2,
  title={Is My Vision-Language Data in Your {AI}? Membership Inference Test ({MINT}) {D}emo 2},
  author={Daniel DeAlcala and others},
  booktitle={IEEE COMPSAC},
year={2026},
}

@inproceedings{pavel25iccv,
  title={{DeepID} Challenge of Detecting Synthetic Manipulations in {ID} Documents},
  author={Pavel Korshunov and others},
  booktitle={IEEE ICCV Workshops},
year={2025},
}

@INPROCEEDINGS { 2023_ECAIw_LFIT-XAI_Tello,
author = {Javier Tello and Marina de la Cruz and Tony Ribeiro and others},
booktitle = {European Conf. on Artificial Intelligence Workshops (ECAIw)},
month = {October},
series = {CEUR-WS},
title = {Symbolic {AI (LFIT) for XAI} to Handle Biases},
volume = {3523},
year = {2023},
}

\vspace{12pt}

\end{document}